\documentclass[11pt]{article}
\usepackage{subcaption}
\usepackage{iftex}
\usepackage[margin=1.5in]{geometry}
\ifPDFTeX
  \usepackage{newtxtext,newtxmath}
\else
  \usepackage{fontspec}
\fi
\usepackage{multirow}
\usepackage{setspace}
\usepackage{enumitem}
\usepackage{tabularx}
\setlist{itemsep=0.15em, topsep=0.25em, parsep=0pt, partopsep=0pt}

\usepackage{amsmath, amssymb, graphicx, booktabs}
\usepackage{xcolor,soul}
\usepackage{svg}
\usepackage{hyperref}
\DeclareRobustCommand{\rev}[1]{{#1}}
\usepackage{pifont}

\title{Latent World Recovery for \rev{Multimodal Learning with Missing Modalities}}
\author{Hui Wang\thanks{These authors contributed equally to this work.},
Tianyu Ren\footnotemark[1], Joseph Butler, Christopher Baker, Karen Rafferty, Simon McDade\\
[0.5em]
Queen's University Belfast, Belfast, United Kingdom
}

\date{}

\begin{document}
\maketitle

\begin{abstract}
We study \rev{multimodal learning under missing modalities}, with particular motivation from bioscience applications in which heterogeneous modalities are often only partially available when decisions need to be made. We propose Latent World Recovery (LWR), a framework built on two key ideas: (i) \rev{modality-specific embeddings} from different modalities are aligned in a shared latent space, and (ii) a unified representation is constructed by fusing only the embeddings of the \rev{modalities} that are actually available at both training and inference time. Rather than imputing missing modalities or requiring a fixed \rev{modality set}, LWR treats each modality as a partial perception of an underlying latent state and performs availability-aware representation learning directly from the observed \rev{modalities}. This combination of neighbor-based latent alignment and \rev{availability-aware modality fusion} enables robust \rev{multimodal prediction} under partial observation, while avoiding error propagation from explicit reconstruction of missing \rev{modalities}. We evaluate the proposed framework on real-world incomplete multi-omics benchmarks and demonstrate that it provides an effective approach to downstream tasks such as cancer phenotype classification and survival prediction.
\end{abstract}
\section{Introduction}
\rev{Multimodal data} are increasingly common in modern bioscience, where the same biological
system can be characterized through heterogeneous modalities such as gene expression,
DNA methylation, copy number variation, protein abundance, imaging, and clinical
variables. Large-scale resources such as The Cancer Genome Atlas have made
such multi-modal molecular profiles widely available for cancer analysis~\cite{tcga2018pancanatlas},
and \rev{multimodal learning} has become an important paradigm for integrating complementary
information across modalities~\cite{ballard2025jasmine,Xing2025.09.15.676314}. In principle, combining
multiple \rev{modalities} can improve predictive modeling, patient stratification, and biological
interpretation. In practice, however, the full set of modalities is rarely available for every
sample. Measurements may be missing because of cost constraints, assay failures,
cohort-specific protocols, or incomplete data integration across studies. This creates a
central challenge for \rev{multimodal learning}: models should exploit \rev{cross-modality} structure when
multiple modalities are observed, while remaining usable when only a subset of \rev{modalities} is
available.

Existing methods address this problem from different perspectives. Early-fusion methods
\cite{beaude2025crossattomics,li2023deep} combine all \rev{modalities} at the input level, but they typically require a fixed and complete \rev{modality}
set. Late-fusion methods \cite{carrillo2022machine,stahlschmidt2022multimodal} combine modality-specific predictions and can naturally operate
with missing \rev{modalities}, but they often capture \rev{cross-modality} interactions only at the decision
level. Shared-representation methods learn a common latent space across modalities, ranging from classical approaches such as canonical correlation analysis (CCA) to deep extensions that use neural networks to capture nonlinear \rev{cross-modality} relationships~\cite{ballard2025jasmine,ma2025moving}. However, many
such methods are most naturally formulated for paired or complete observations. Generative
approaches, particularly multimodal variational autoencoders, provide a principled
latent-variable formulation for incomplete multimodal learning.
They can model shared latent variables and support inference under missing modalities,
but they often emphasize reconstruction or cross-modal generation. In multi-omics
applications, recent methods such as MIND, JASMINE, and IntegrAO have made important
progress in learning representations from incomplete omics profiles through VAE-based,
self-supervised, or graph-based integration strategies~\cite{Xing2025.09.15.676314,ma2025moving,ballard2025jasmine}.
Nevertheless, a persistent tension remains between two desiderata: learning a coherent
shared latent organization across modalities and making downstream decisions directly from
whatever \rev{modalities} are actually observed.

In this work, we take a representation-learning perspective on incomplete \rev{multimodal}
learning. Rather than treating missing modalities as targets that must be synthesized before
downstream analysis, we view each available modality as a partial observation of an
underlying latent biological state. Under this \rev{perspective}, the goal is not to reconstruct all
modalities, but to recover a useful sample-level representation from the observed \rev{modalities}.
This distinction is important in multi-omics data, where different assays may reflect related
but non-identical biological processes. Forcing one modality to exactly reproduce another
can obscure modality-specific variation, while ignoring \rev{cross-modality} structure can lead to
fragmented representations.

We propose \emph{Latent World Recovery} (LWR), a VAE-based framework for \rev{multimodal}
representation learning with missing \rev{modalities}. LWR has three main components. First, each
modality is encoded by a \rev{modality-specific variational encoder}, while all encoders map their
inputs into a shared latent space. Second, for each sample, LWR constructs a fused
representation by aggregating only the embeddings of the modalities that are observed.
Missing \rev{modalities} are not replaced by zero vectors, mask tokens, or imputed features; they are
excluded from the fusion operation. Third, instead of enforcing strict coordinate-level
agreement between modality embeddings, LWR introduces a neighbor-based alignment
objective. This objective encourages the fused representation to preserve the local
sample-neighborhood structures induced by individual modalities, thereby aligning
\rev{cross-modality} relational information without requiring different modalities to collapse to
identical latent coordinates.

Importantly, LWR does not synthesize missing modalities as an intermediate step for
downstream prediction. It uses reconstruction of observed \rev{modalities} as a self-supervised
training signal, together with variational regularization and neighbor-based latent alignment.
After training, the fused latent representation is extracted as a general-purpose sample
embedding and evaluated using external downstream models. This separates \rev{incomplete-modality}
representation learning from task-specific prediction, allowing the same learned
representation to be used for cancer phenotype classification, survival prediction, reconstruction-
based information preservation, and clustering-based patient stratification.

We evaluate LWR on incomplete multi-omics benchmarks derived from TCGA \cite{tcga2012}, CCMA \cite{sun2023generation}, and
CCLE \cite{Ghandi2019}. Across cancer phenotype classification, survival prediction, and reconstruction analysis,
LWR achieves competitive or superior performance compared with representative
multi-omics integration methods. We further conduct ablation studies to examine the
effects of fusion and alignment strategies. The results show that naive pairwise alignment
can substantially degrade the learned representations, whereas neighbor-based alignment
provides a more stable way to preserve cross-sample structure. A clustering-based survival
stratification case study further suggests that the learned latent space captures clinically
meaningful patient heterogeneity.

Our contributions are summarized as follows:
\begin{itemize}
    \item We formulate incomplete \rev{multimodal learning} as latent-state recovery from partial
    observations, shifting the focus from \rev{missing-modality synthesis} to representation learning
    from available \rev{modalities}.
    \item We propose LWR, a VAE-based framework that combines availability-aware latent
    fusion with neighbor-based alignment of modality-induced sample structures.
    \item We evaluate LWR on incomplete multi-omics benchmarks across cancer phenotype
    classification, survival prediction, reconstruction analysis, ablation studies, and
    clustering-based survival stratification.
\end{itemize}

\section{Related Work}
Learning shared representations from multiple \rev{modalities} has a long history in \rev{the broader multiview learning literature}. CCA seeks linear projections that maximize \rev{cross-modality} correlation and remains a foundational reference for shared-latent-space learning \cite{Hotelling1936}. Deep CCA extends this idea to nonlinear feature mappings learned by neural networks, making correlation-based alignment compatible with modern high-capacity encoders \cite{Andrew2013}. More broadly, multimodal and multiview representation learning has been organized around alignment, fusion, and co-learning principles in influential surveys, which provide the right conceptual framing for methods that must combine agreement across modalities with complementary modality-specific information \cite{Baltrusaitis2019,Li2018}. In this sense, LWR is best viewed not simply as another fusion architecture, but as a method that jointly addresses latent alignment and robustness to partial observation.

\paragraph{Contrastive Objectives and Modality Incompleteness.} Modern multimodal representation learning has further emphasized large-scale representation alignment using contrastive or correspondence-based objectives. Representative works such as CLIP demonstrate that cross-modal agreement in a shared embedding space can produce highly transferable representations, while self-supervised multimodal pretraining frameworks such as VATT show that such alignment can be learned even from weak supervision across heterogeneous signals \cite{Radford2021,Akbari2021}. These methods are highly influential because they establish the effectiveness of discriminative alignment objectives; however, they typically assume abundant paired training data and are not designed for the structured, sample-wise incompleteness that is common in biomedical multi-omics cohorts. This distinction is important for our setting, where the central challenge is not merely cross-modal transfer, but learning stable patient representations when any subset of modalities may be absent.

\paragraph{Generative Modeling.} A second major line of work treats multimodal learning from a generative perspective. Variational autoencoders provide the basic latent-variable framework for learning stochastic encoders and decoders from high-dimensional observations \cite{Kingma2014}. Building on this foundation, MVAE introduced a product-of-experts inference mechanism that can naturally handle arbitrary subsets of observed modalities, and MMVAE proposed a mixture-of-experts formulation that improves cross-generation and shared/private factor modeling in multimodal generative learning \cite{Wu2018,Shi2019}. In parallel, masked modeling has become a strong paradigm for representation learning from partial observations, with MAE showing that reconstructing masked inputs can yield scalable and transferable representations \cite{He2022}. Closely related ideas also appear in mixture-of-experts and gating models, where conditional routing is used to adapt computation or information aggregation to the input; classical deep MoE and later multi-gate MoE formulations are particularly relevant as references for availability-aware or selectively weighted fusion \cite{Eigen2014,Ma2018}. Compared with these approaches, LWR is not a pure generative missing-modality model nor a pure routing-based fusion model: it combines \rev{modality-specific stochastic encoders}, \rev{observed-modality reconstruction}, neighborhood-preserving latent alignment, and adaptive fusion in a representation-learning objective tailored to incomplete multi-omics data.

\paragraph{Multi-Omics Integration and Relational Topology.} Within computational biology, multi-omics integration has increasingly moved from early concatenation toward latent, graph, and relation-aware integration. TCGA and CCLE have become canonical resources for evaluating whether learned representations capture clinically or biologically meaningful structure across heterogeneous molecular measurements \cite{Weinstein2013,Ghandi2019}. Graph-based integration is especially relevant here, because patient-patient or sample-sample relationships often remain informative even when some omics layers are only partially overlapping; similarity network fusion and network embedding methods provide useful precedents for preserving relational topology rather than relying solely on feature-level agreement \cite{Wang2014SNF,Grover2016}. Recent accessible deep-learning methods for incomplete or heterogeneous multi-omics include IntegrAO \cite{ma2025moving}, which fuses partially overlapping patient graphs with graph neural networks; CLCLSA \cite{Zhao2023CLCLSA}, which combines cross-omics autoencoding with self-attention; and OmiEmbed \cite{Zhang2021OmiEmbed}, which learns unified omics embeddings in a multi-task framework. 

Closely related to our problem setting is MIND \cite{Xing2025.09.15.676314}, which addresses incomplete multi-omics integration through a multimodal variational autoencoder framework. MIND aggregates available modality-specific embeddings via unweighted averaging and explicitly predicts unobserved missing modalities from the fused representation. To preserve relational structure, MIND regularizes the latent space using a static, exponential-tilted prior derived from input-space t-SNE affinity matrices, alongside a coordinate-wise mean squared error penalty that forces representations of different modalities from the same patient to be identical.

Relative to these methods, and MIND in particular, LWR introduces fundamental structural departures to avoid over-regularization and error propagation. First, rather than relying on uniform averaging, LWR employs an availability-aware attention mechanism to dynamically weight the contribution of each observed modality. Second, LWR strictly bounds its reconstruction objective to observed modalities, avoiding the synthesis of noise that can arise from explicitly predicting missing omics features. By replacing static, input-space priors with a dynamic stop-gradient KL divergence loss, LWR matches the fused representation directly to the continuously learned neighborhood topologies of the modality-specific latent spaces.
\section{Method}
\begin{figure}
    \centering
    \includegraphics[width=0.9\textwidth]{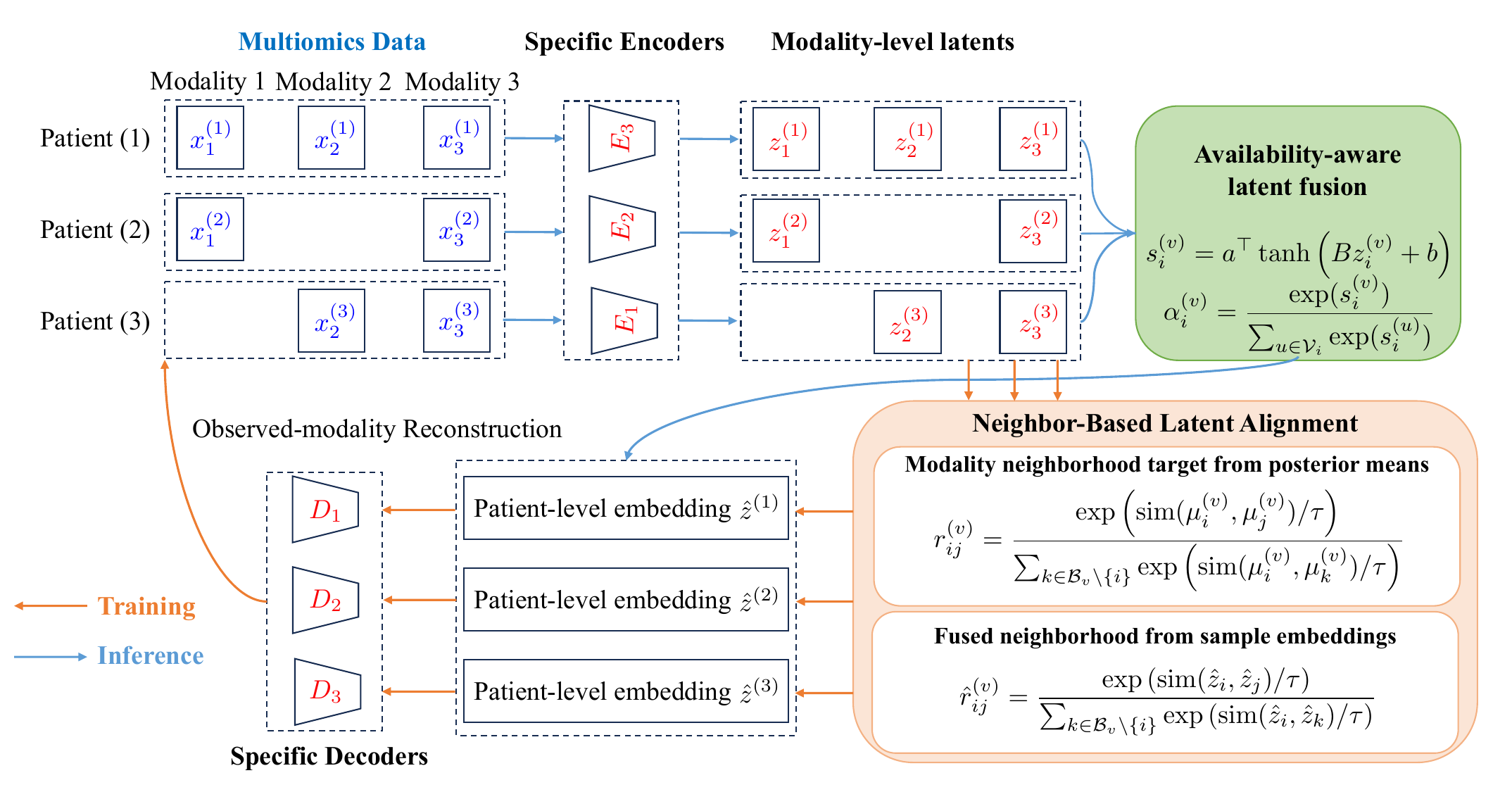}
    \caption{Overview of Latent World Recovery (LWR). Each available modality is encoded by a \rev{modality-specific variational encoder} into a shared latent space. The sample-level representation is obtained by fusing only the observed \rev{modality embeddings} through \rev{availability-aware modality fusion}. During training, LWR reconstructs observed \rev{modalities} and applies neighbor-based latent alignment to preserve modality-induced local sample structures. At inference time, missing \rev{modalities} are excluded rather than imputed.}
    \label{fig:1}
\end{figure}
\subsection{Problem Setup}
We consider a \rev{multimodal dataset} with $N$ samples and $M$ possible \rev{modalities}. Let $\mathcal{V}=\{1,\ldots,M\}$ denote the complete set of \rev{modalities}. For sample $i$, only a subset of \rev{modalities} $\mathcal{V}_i \subseteq \mathcal{V}$ is observed. The input of sample $i$ is therefore
\begin{equation}
    \mathbf{x}_i = \{x_i^{(v)}: v \in \mathcal{V}_i\},
\end{equation}
where $x_i^{(v)} \in \mathbb{R}^{d_v}$ denotes the feature vector of \rev{modality} $v$. The \rev{observed-modality set} $\mathcal{V}_i$ may vary across samples, reflecting the \rev{incomplete-modality setting}.

We assume that heterogeneous \rev{modalities} provide partial observations of an underlying latent biological state. The goal of Latent World Recovery (LWR) is to learn a unified sample-level representation from the available \rev{modalities} without requiring a complete set of modalities (see Figure~\ref{fig:1} for an overview of the pipeline). Rather than reconstructing missing modalities as an intermediate step for downstream analysis, LWR directly aggregates the latent information provided by the observed \rev{modalities}. The learned representation can then be used by external downstream models for tasks such as cancer phenotype classification and survival prediction.
\subsection{\rev{Modality-Specific Encoding} and Shared-Latent Projection}
For each \rev{modality} $v$, LWR uses a \rev{modality-specific variational encoder} (VAE) \cite{Kingma2014} to map the corresponding input into a shared latent space. Given an observed \rev{modality} $x_i^{(v)}$, the encoder produces the parameters of a Gaussian posterior:
\begin{equation}
    q_{\phi_v}(z_i^{(v)} \mid x_i^{(v)})=
    \mathcal{N}\left(\mu_i^{(v)}, \operatorname{diag}\left((\sigma_i^{(v)})^2\right)
    \right),
\end{equation}
where $\mu_i^{(v)}, \sigma_i^{(v)} \in \mathbb{R}^{d_z}$. The latent variable is sampled using the reparameterization trick:
\begin{equation}
    z_i^{(v)}=\mu_i^{(v)}+\sigma_i^{(v)} \odot \epsilon, \qquad \epsilon \sim \mathcal{N}(0,I).
\end{equation}
This formulation allows each modality to have its own encoder while mapping all modalities into a common latent space. In practice, the posterior means $\mu_i^{(v)}$ are also used as deterministic \rev{modality-level representations} when constructing neighborhood relations.

\subsection{Availability-Aware Latent Fusion}
Given the latent representations of the observed \rev{modalities}, LWR constructs a sample-level  embedding by fusing only the available modalities. For sample $i$, the fusion module takes $\{z_i^{(v)}:v\in\mathcal{V}_i\}$ as input and computes an adaptive weight for each observed \rev{modality}:
\begin{equation}
    s_i^{(v)}=a^\top \tanh \left(B z_i^{(v)} + b\right),\qquad v \in \mathcal{V}_i,
\end{equation}
where $a$, $B$, and $b$ are learnable parameters. The weights are normalized over the observed \rev{modalities} only:
\begin{equation}
\alpha_i^{(v)}=\frac{\exp(s_i^{(v)})}{\sum_{u\in\mathcal{V}_i}\exp(s_i^{(u)})},\qquad v \in \mathcal{V}_i .
\end{equation}
The fused latent representation is then defined as
\begin{equation}
\hat{z}_i =\sum_{v\in\mathcal{V}_i}\alpha_i^{(v)} z_i^{(v)} .
\end{equation}
This fusion rule is availability-aware by construction. Missing \rev{modalities} are not replaced by imputed features, zero vectors, or learned mask tokens; they are simply excluded from the aggregation. The same fusion mechanism is used during training and inference, so the model is optimized under the same \rev{incomplete-modality} condition in which it is later applied.
\subsection{\rev{Observed-Modality Reconstruction}}
Although LWR does not reconstruct missing modalities as a prerequisite for downstream prediction, we use reconstruction of observed \rev{modalities} as a self-supervised training signal. Specifically, the available \rev{modality-specific latent variables} are first fused into a shared latent representation \(\hat{z}_i\). For each \rev{modality} \(v\), a \rev{modality-specific decoder} reconstructs the corresponding input from this fused representation:
\begin{equation}
\hat{x}_i^{(v)} = D_v(\hat{z}_i),
\end{equation}
where \(D_v\) denotes the decoder for \rev{modality} \(v\). Importantly, the reconstruction objective is evaluated only for observed \rev{modalities}; missing \rev{modalities} are not reconstructed or used in the loss. Let \(m_i^{(v)} \in \{0,1\}\) indicate whether \rev{modality} \(v\) is observed for sample \(i\), and let \(d_v\) be the dimensionality of \rev{modality} \(v\). The reconstruction loss is
\begin{equation}
\mathcal{L}_{\mathrm{rec}}
=
\frac{1}{|\mathcal{V}|}
\sum_{v\in\mathcal{V}}
\frac{
\sum_{i=1}^{N}
m_i^{(v)}
\left\|
x_i^{(v)} - \hat{x}_i^{(v)}
\right\|_2^2
}{
d_v \sum_{i=1}^{N} m_i^{(v)} + \epsilon
},
\end{equation}
where \(\epsilon\) is a small constant for numerical stability. For continuous omics features, this corresponds to a masked mean squared error over observed sample--\rev{modality} pairs. This objective encourages the fused latent representation to preserve information from the available \rev{modalities}, while avoiding the stronger requirement that unobserved \rev{modalities} must be synthesized before downstream analysis.

The variational regularization term is computed only for observed \rev{modalities}:
\begin{equation}
\mathcal{L}_{\mathrm{KL}}
=
\frac{1}{|\mathcal{V}|}
\sum_{v\in\mathcal{V}}
\frac{
\sum_{i=1}^{N}
m_i^{(v)}
D_{\mathrm{KL}}
\left(
q_{\phi_v}(z_i^{(v)} \mid x_i^{(v)})
\,\|\, \mathcal{N}(0,I)
\right)
}{
\sum_{i=1}^{N} m_i^{(v)} + \epsilon
}.
\end{equation}
This term regularizes the \rev{modality-specific posteriors} and stabilizes the shared latent space without imposing any penalty on missing \rev{modalities}.

\subsection{Neighbor-Based Latent Alignment}
The fused representation should not only aggregate information from available \rev{modalities}, but also preserve the sample relationships captured by individual modalities. In multi-omics data, each modality may induce a meaningful local structure over samples, such as grouping biologically similar patients closer in the latent space. We therefore introduce a neighbor-based alignment objective that encourages the fused representation to preserve these modality-specific neighborhood structures.

Let $\mathcal{B}$ denote a mini-batch, and let
$ \mathcal{B}_v=\{i\in\mathcal{B}: m_i^{(v)}=1\}$ be the subset of samples in the mini-batch for which modality $v$ is observed. For each observed modality $v$, we compute a modality-induced neighborhood distribution only over samples in $\mathcal{B}_v$. For an anchor sample $i\in\mathcal{B}_v$, the neighborhood probability assigned to another sample $j\in\mathcal{B}_v \setminus \{i\}$ is defined as
\begin{equation}
r_{ij}^{(v)}=
\frac{
\exp\left(\operatorname{sim}(\mu_i^{(v)},\mu_j^{(v)})/\tau\right)
}{
\sum_{k\in\mathcal{B}_v \setminus \{i\}}
\exp\left(\operatorname{sim}(\mu_i^{(v)},\mu_k^{(v)})/\tau\right)
},
\qquad j\in\mathcal{B}_v \setminus \{i\},
\end{equation}
where $\operatorname{sim}(\cdot,\cdot)$ denotes cosine similarity and $\tau$ is a temperature parameter controlling the sharpness of the neighborhood distribution. A smaller $\tau$ places more emphasis on the nearest samples, whereas a larger $\tau$ produces a smoother distribution over the modality-specific mini-batch neighborhood.

To compare this modality-induced structure with the fused latent space, we compute the corresponding fused neighborhood distribution over the same support:
\begin{equation}
\hat r_{ij}^{(v)}
=
\frac{
\exp\left(\operatorname{sim}(\hat z_i,\hat z_j)/\tau\right)
}{
\sum_{k\in\mathcal{B}_v \setminus \{i\}}
\exp\left(\operatorname{sim}(\hat z_i,\hat z_k)/\tau\right)
},
\qquad j\in\mathcal{B}_v \setminus \{i\}.
\end{equation}
Here, the superscript $(v)$ in $\hat r_{ij}^{(v)}$ indicates that the fused-space neighborhood distribution is restricted to the same sample subset $\mathcal{B}_v$ used by modality $v$. This ensures that the modality-specific and fused distributions are defined over identical supports.

The neighbor-based alignment loss is then defined as
\begin{equation}
\mathcal{L}_{\mathrm{neigh}}=
\frac{1}{|\mathcal{A}|}
\sum_{(i,v)\in\mathcal{A}}
D_{\mathrm{KL}}
\left(
\operatorname{sg}(r_i^{(v)})\parallel
\hat r_i^{(v)}
\right),
\end{equation}
where $\mathcal{A}=\{(i,v): i\in\mathcal{B}_v,\ |\mathcal{B}_v|>1\}$.
Here, $r_i^{(v)}$ and $\hat r_i^{(v)}$ denote the neighborhood distributions of anchor sample $i$ under modality $v$ and under the fused representation, respectively. The operator $\operatorname{sg}(\cdot)$ denotes stop-gradient. Thus, the modality-specific neighborhood distribution is treated as a structural target, while the fused representation is optimized to match it.

This stop-gradient design stabilizes training by preventing both sides of the alignment objective from being updated simultaneously. It also makes the role of the alignment loss explicit: the objective transfers local sample structure from each observed modality to the fused latent representation. Compared with coordinate-level matching between modality embeddings, this relational formulation constrains the induced sample neighborhoods rather than requiring exact agreement of individual latent coordinates. As a result, LWR encourages a coherent shared latent organization while allowing each modality to retain modality-specific variation.
\subsection{Overall Training Objective}
The final LWR objective combines \rev{observed-modality reconstruction}, variational regularization, and neighbor-based latent alignment:
\begin{equation}
    \mathcal{L} = \lambda_{\mathrm{rec}}\mathcal{L}_{\mathrm{rec}}
    +
    \lambda_{\mathrm{KL}}\mathcal{L}_{\mathrm{KL}}
+\lambda_{\mathrm{neigh}}\mathcal{L}_{\mathrm{neigh}} .
\end{equation}

Here, $\lambda_{\mathrm{rec}}$, $\lambda_{\mathrm{KL}}$, and $\lambda_{\mathrm{neigh}}$ control the relative contributions of the reconstruction, variational, and neighborhood alignment terms. No task-specific prediction head is used during representation learning. After training, the fused latent representation $\hat{z}_i$ is extracted as the sample-level embedding and evaluated using separate downstream models. This separates representation learning from task-specific prediction and allows the same learned embedding to be used across various downstream tasks.

\subsection{Discussion of the Method}
LWR can be viewed as a VAE-based incomplete \rev{multimodal representation learning framework} with \rev{availability-aware modality fusion} and neighbor-preserving latent alignment. Its key distinction is that missing \rev{modalities} are not treated as targets that must be explicitly synthesized before downstream analysis. Instead, each available \rev{modality} contributes a partial latent estimate, and the final representation is recovered by fusing only the observed modalities.

The neighbor-based alignment objective further distinguishes LWR from strict coordinate-level alignment. Rather than enforcing different modalities to produce identical embeddings for the same sample, it aligns their induced local sample structures. This is important for multi-omics learning, where modalities can be biologically related while still carrying \rev{modality-specific signals}. As a result, LWR encourages a shared latent organization without eliminating modality-specific variation.
\section{Experiments}
\subsection{Experimental Setup}
\paragraph{Datasets.}
We conduct experiments on the same multi-omics benchmarks used in \cite{Xing2025.09.15.676314}, including (1) \textbf{The Cancer Genome Atlas} (\textbf{TCGA}), (2) \textbf{Childhood Cancer Model Atlas} (\textbf{CCMA}), and (3) \textbf{Cancer Cell Line Encyclopedia} (\textbf{CCLE}). For TCGA, we use 17 cancer cohorts: LUAD, KIRC, BRCA, LGG, OV, SKCM, THCA, BLCA, STAD, UCEC, COADREAD, COAD, GBM, HNSC, LUSC, LIHC, and CESC. Each cohort contains up to five omics modalities, including mRNA-seq, miRNA-seq, DNA methylation, copy number variation (CNV), and reverse phase protein array (RPPA), together with available clinical annotations. CCMA consists of 182 pediatric cancer cell lines profiled with mRNA-seq, DNA methylation, and CNV modalities, while CCLE contains 1461 cancer cell lines profiled with RNA-seq, DNA methylation, CNV, microRNA, RPPA, and metabolomics. These datasets span diverse biological contexts and contain naturally incomplete modality availability, making them suitable for evaluating \rev{multimodal representation learning} under realistic missing-modality patterns.

\paragraph{Downstream Tasks.}
We evaluate the learned representations on three downstream tasks: \textbf{cancer phenotype classification}, \textbf{survival prediction}, and \textbf{data reconstruction}. For classification, we train XGBoost classifiers \cite{Chen_2016} on the learned embeddings and report five-fold cross-validation accuracy. Specifically, the classification task corresponds to tumor stage prediction for TCGA cohorts and cancer type prediction for CCMA and CCLE. For survival prediction, we train a penalized Cox proportional hazards model \cite{simon2011regularization} using the learned embeddings together with available clinical covariates, and report the concordance index (C-index) \cite{harrell1996multivariable} under five-fold cross-validation. This task is evaluated on TCGA cohorts with available survival annotations. For data reconstruction, we assess whether the learned representations preserve modality-specific information by reconstructing masked omics values and measuring the Pearson correlation between the reconstructed and observed values. \rev{Specifically, following MIND}~\cite{Xing2025.09.15.676314}, \rev{we mask 10\% of originally observed feature values within each modality before training and evaluate reconstruction only on these held-out values, not on naturally missing entries, held-out samples, or entirely held-out modalities.} These tasks jointly evaluate the predictive utility and information-preserving capability of the learned representations under incomplete multi-omics settings.
\paragraph{Baselines.} We compare LWR with four state-of-the-art multi-omics integration methods:
\begin{itemize}
\item \textbf{MIND}~\cite{Xing2025.09.15.676314}: a multimodal VAE for incomplete multi-omics integration. It uses modality-specific encoders to obtain individual omics embeddings, aggregates available embeddings into a patient-level representation, and introduces a neighborhood-aware prior based on modality-specific affinity structures to preserve biological sample relationships.
\item \textbf{IntegrAO}~\cite{ma2025moving}: a graph-based method for integrating arbitrary combinations of omics modalities. It constructs modality-specific sample similarity networks and combines them to learn unified patient representations without requiring complete modality overlap.

\item \textbf{JASMINE}~\cite{ballard2025jasmine}: a VAE-based framework for incomplete multi-omics representation learning. It models both modality-specific and shared latent information through multiple interacting encoders and decoders, enabling cross-modal information sharing under missing-modality settings.

\item \textbf{MSNE}~\cite{xu2021network}: a network embedding method for partial multi-omics integration. It learns low-dimensional sample representations by preserving graph-derived relationships across omics \rev{modalities}.
\end{itemize}

\rev{To ensure strict comparability, our downstream evaluation rigorously follows the standardized benchmarking protocol established by MIND}~\cite{Xing2025.09.15.676314}. \rev{The performance metrics for all baseline methods are directly taken from the reported results in the MIND study, which were evaluated under the exact same data splits, incomplete-modality settings, and downstream predictive models. Because these baseline metrics are extracted directly from the literature, we report their provided point estimates in our comparisons. For reconstruction, we report only methods with native decoder-based reconstruction outputs under the MIND protocol. IntegrAO and MSNE are omitted from Table}~\ref{tab:reconstruction} \rev{because they do not include modality-specific decoders or a reconstruction module; adding an external decoder would introduce an extra model component and would not be directly comparable to their original formulations.}

\paragraph{Implementation Details.} We follow the same preprocessing pipeline as \cite{Xing2025.09.15.676314}. For TCGA, mRNA expression, DNA methylation, microRNA, protein expression, and clinical annotations are obtained from Broad Firehose, while CNV is obtained from cBioPortal. Features with more than 20\% missing values are removed within each modality, and the remaining missing values are imputed using k-nearest neighbors. Following the original benchmark, the top 2,000 most variable features are selected for high-dimensional modalities, including CNV, mRNA/RNA-seq, and DNA methylation, where applicable. All modalities are then standardized to zero mean and unit variance. The same preprocessing protocol is adopted for CCMA and CCLE following \cite{Xing2025.09.15.676314}. The main training hyperparameters and LWR configurations are summarized in Table~\ref{tab:lwr_implementation}.
\begin{table}[h]
\centering
\begin{tabular}{lc}
\toprule
Parameter & Value \\
\midrule
Encoder hidden dimension & 512 \\
Latent dimension & 64 \\
Dropout rate & 0.2 \\
Optimizer & AdamW \\
Learning rate & $1 \times 10^{-3}$ \\
Weight decay & $1 \times 10^{-4}$ \\
Batch size & 32 \\
Maximum epochs & 300 \\
Validation fraction & 0.15 \\
Early stopping patience & 50 \\
Reconstruction loss weight & 1.0 \\
KL loss weight  & $1 \times 10^{-3}$ \\
Alignment loss weight & 0.03 \\
Neighborhood temperature & 0.2 \\
\bottomrule
\end{tabular}
\caption{Implementation details and hyperparameters of LWR.}
\label{tab:lwr_implementation}

\end{table}
\subsection{Main Results}
\begin{table}[t]
\centering
\small
\setlength{\tabcolsep}{3pt}
\setlength{\extrarowheight}{1.2pt}
\begin{tabular}{c|cccccc}
\toprule
Dataset & Cancer Type & MIND \cite{Xing2025.09.15.676314} & IntegrAO \cite{ma2025moving} & JASMINE \cite{ballard2025jasmine} & MSNE \cite{xu2021network} & \textbf{LWR (ours)} \\
\hline
\multirow{17}{*}{TCGA}
& LUAD     & \textbf{0.517 (1)} & 0.476 (4) & \underline{0.505 (2)} & 0.474 (5) & 0.500 (3) \\
& KIRC     & 0.513 (3) & \underline{0.552 (2)} & 0.510 (4) & 0.460 (5) & \textbf{0.557 (1)} \\
& BRCA     & 0.516 (4) & 0.506 (5) & \underline{0.540 (2)} & \textbf{0.545 (1)} & 0.537 (3) \\
& LGG      & - & - & - & - & - \\
& OV       & - & - & - & - & - \\
& SKCM     & \underline{0.456 (2)} & 0.443 (4) & \textbf{0.491 (1)} & 0.438 (5) & 0.450 (3) \\
& THCA     & \underline{0.543 (2)} & 0.540 (3) & \textbf{0.573 (1)} & 0.533 (4) & 0.505 (5) \\
& BLCA     & \underline{0.451 (2)} & 0.417 (4) & 0.432 (3) & 0.337 (5) & \textbf{0.468 (1)} \\
& STAD     & 0.383 (3) & 0.377 (4) & \underline{0.425 (2)} & 0.376 (5) & \textbf{0.456 (1)} \\
& UCEC     & - & - & - & - & - \\
& COADREAD & \underline{0.333 (2)} & 0.282 (5) & 0.331 (3) & 0.324 (4) & \textbf{0.335 (1)} \\
& COAD     & 0.334 (4) & 0.324 (5) & \underline{0.351 (2)} & \textbf{0.358 (1)} & 0.335 (3) \\
& GBM      & - & - & - & - & - \\
& HNSC     & \underline{0.541 (2)} & 0.496 (5) & 0.529 (3) & 0.501 (4) & \textbf{0.555 (1)} \\
& LUSC     & \underline{0.472 (2)} & 0.442 (4) & 0.470 (3) & 0.430 (5) & \textbf{0.488 (1)} \\
& LIHC     & \textbf{0.480 (1)} & \underline{0.473 (2)} & 0.472 (3) & 0.371 (5) & 0.464 (4) \\
& CESC     & 0.453 (3) & 0.427 (5) & \textbf{0.503 (1)} & 0.447 (4) & \underline{0.480 (2)} \\ \hline
CCMA & CCMA & \underline{0.793 (2)} & 0.761 (3) & 0.704 (4) & 0.548 (5) & \textbf{0.844 (1)} \\ \hline
CCLE & CCLE & \textbf{0.659 (1)} & 0.547 (4) & \underline{0.610 (2)} & 0.367 (5) & 0.577 (3) \\
\hline
\multicolumn{2}{c}{Average rank} & 2.27 & 3.93 & 2.40 & 4.20 & \textbf{2.20} \\
\bottomrule
\end{tabular}
\caption{Five-fold cross-validation accuracy for classification on \textbf{TCGA}, \textbf{CCMA}, and \textbf{CCLE} datasets. Following \cite{Xing2025.09.15.676314}, we use XGBoost as the downstream classifier, taking the embeddings produced by each method as input. For TCGA cohorts, the task is tumor stage classification; for CCMA and CCLE, the task is cancer type classification. The best result on each dataset is highlighted in bold, and the second-best result is underlined. Numbers in parentheses indicate the rank of each method on the corresponding dataset.}
\label{tab:method_comparison}
\end{table}

\begin{table}[t]
\centering
\small
\begin{tabular}{cccccc}
\toprule
Dataset/Method & MIND \cite{Xing2025.09.15.676314} & IntegrAO \cite{ma2025moving} & JASMINE \cite{ballard2025jasmine} & MSNE \cite{xu2021network} & \textbf{LWR (ours)} \\
\midrule
LUAD & \textbf{0.572 (1)} & 0.559 (3) & 0.553 (4) & 0.474 (5) & \underline{0.562 (2)} \\
KIRC & \textbf{0.724 (1)} & 0.691 (4) & 0.705 (3) & 0.591 (5) & \underline{0.706 (2)} \\
BRCA & \underline{0.650 (2)} & 0.646 (3) & 0.640 (4) & 0.634 (5) & \textbf{0.671 (1)} \\
LGG & \textbf{0.836 (1)} & \underline{0.828 (2)} & 0.809 (4) & 0.746 (5) & 0.812 (3) \\
OV & \textbf{0.609 (1)} & \underline{0.601 (2)} & 0.593 (5) & 0.596 (4) & 0.598 (3) \\
SKCM & \textbf{0.646 (1)} & 0.599 (5) & \underline{0.633 (2)} & 0.601 (4) & 0.614 (3) \\
THCA & 0.894 (3) & \textbf{0.932 (1)} & 0.711 (5) & \underline{0.921 (2)} & 0.760 (4) \\
BLCA & \underline{0.622 (2)} & 0.592 (5) & \textbf{0.630 (1)} & 0.598 (4) & 0.615 (3) \\
STAD & \underline{0.555 (2)} & 0.553 (3) & 0.526 (5) & 0.535 (4) & \textbf{0.563 (1)} \\
UCEC & 0.646 (3) & \underline{0.664 (2)} & 0.582 (4) & 0.579 (5) & \textbf{0.665 (1)} \\
COADREAD & 0.564 (3) & \underline{0.602 (2)} & 0.540 (5) & 0.558 (4) & \textbf{0.608 (1)} \\
COAD & 0.540 (4) & \underline{0.553 (2)} & 0.531 (5) & \textbf{0.573 (1)} & 0.541 (3) \\
GBM & 0.647 (3) & \textbf{0.654 (1)} & 0.634 (5) & 0.642 (4) & \underline{0.651 (2)} \\
HNSC & \textbf{0.612 (1)} & \underline{0.586 (2)} & 0.572 (4) & 0.546 (5) & 0.577 (3) \\
LUSC & \underline{0.542 (2)} & 0.531 (4) & 0.540 (3) & \textbf{0.558 (1)} & 0.495 (5) \\
LIHC & \underline{0.606 (2)} & 0.569 (4) & 0.593 (3) & 0.514 (5) & \textbf{0.639 (1)} \\
CESC & 0.644 (3) & 0.594 (4) & \textbf{0.673 (1)} & 0.560 (5) & \underline{0.651 (2)} \\
\midrule
Average rank & \textbf{2.06} & 2.88 & 3.71 & 4.00 & \underline{2.35} \\
\bottomrule
\end{tabular}
\caption{Survival prediction on \textbf{TCGA} datasets. Following \cite{Xing2025.09.15.676314}, we use penalized Cox’s proportional hazards models as the downstream predictor, taking patient information and the embeddings produced by each method as input. The C-index is estimated using five-fold cross-validation. The best result on each dataset is highlighted in bold, and the second-best result is underlined. Numbers in parentheses indicate the rank of each method on the corresponding dataset, with smaller ranks indicating better performance. The average rank is computed over datasets for which results are available.}
\label{tab:survival_prediction}
\end{table}

\begin{table}[htb]
\centering
\small
\setlength{\extrarowheight}{1.2pt}
\begin{tabular}{c|cccc}
\toprule
Dataset & Cancer Type & MIND \cite{Xing2025.09.15.676314} & JASMINE \cite{ballard2025jasmine} & \textbf{LWR (ours)} \\
\midrule
\multirow{17}{*}{TCGA}  
&LUAD     & \underline{0.275} & 0.174 & \textbf{0.365} \\
&KIRC     & \underline{0.393}  & 0.186 & \textbf{0.475} \\
&BRCA     & \underline{0.378}  & 0.169 & \textbf{0.473} \\
&LGG      & \underline{0.499}  & 0.250 & \textbf{0.591} \\
&OV       & \underline{0.205} & 0.183 & \textbf{0.331} \\
&SKCM     & \underline{0.283}  & 0.189 & \textbf{0.412} \\
&THCA     & \underline{0.309}  & 0.237 & \textbf{0.453} \\
&BLCA     & \underline{0.340}  & 0.212 & \textbf{0.436} \\
&STAD     & \underline{0.352} & 0.086 & \textbf{0.436} \\
&UCEC     & \underline{0.326}  & 0.236 & \textbf{0.459} \\
&COADREAD & \underline{0.308}  & 0.168 & \textbf{0.443} \\
&COAD     & \underline{0.264} & 0.173 & \textbf{0.368} \\
&GBM      & \underline{0.203}  & 0.158 & \textbf{0.267} \\
&HNSC     & \underline{0.258}  & 0.169 & \textbf{0.399} \\
&LUSC     & \underline{0.265}  & 0.176 & \textbf{0.355} \\
&LIHC     & \underline{0.417} & 0.192 & \textbf{0.527} \\
&CESC     & \underline{0.270} & 0.199 & \textbf{0.332} \\ \hline
CCMA&CCMA     & \underline{0.678} & 0.362 & \textbf{0.761} \\ \hline
CCLE&CCLE     & \underline{0.359} & 0.180 & \textbf{0.410} \\
\bottomrule
\end{tabular}
\caption{Reconstruction performance on \textbf{TCGA, CCMA} and \textbf{CCLE} datasets. \rev{Following MIND}~\cite{Xing2025.09.15.676314}, \rev{10\% of originally observed feature values are masked within each modality before training and evaluated as held-out reconstruction targets.} Pearson correlations between prediction and observed values are averaged over all modalities. Here, JASMINE is augmented with 64-dimensional modality-specific embeddings. The best result on each cancer type is highlighted in bold, and the second-best result is underlined.}
\label{tab:reconstruction}
\end{table}
Tables \ref{tab:method_comparison}-\ref{tab:reconstruction} report the main experimental results on cancer phenotype classification, survival prediction, and data reconstruction. Overall, LWR achieves competitive or superior performance across the evaluated multi-omics benchmarks. Unlike methods that require complete modality inputs or explicitly impute missing modalities before representation learning, LWR produces unified embeddings by aggregating only the available omics \rev{modalities} in the shared latent space. This demonstrates that \rev{availability-aware modality fusion} can learn effective sample-level representations under incomplete multi-omics settings.

For cancer phenotype classification (Table~\ref{tab:method_comparison}), LWR obtains the best average rank among all compared methods, with an average rank of 2.20 across TCGA, CCMA, and CCLE. On TCGA, LWR achieves the best performance on multiple cancer cohorts, including KIRC, BLCA, STAD, COADREAD, HNSC, and LUSC. It also obtains the best result on CCMA, improving the accuracy from 0.793 for the strongest baseline to 0.844. These results indicate that the representations learned by LWR are highly discriminative for downstream classification. Although LWR is not the best-performing method on every individual cohort, its consistently strong ranking across datasets suggests that learning a shared latent representation from available modalities provides a robust strategy for cancer phenotype prediction.

For survival prediction (Table~\ref{tab:survival_prediction}), LWR also shows strong performance on TCGA datasets. It achieves the best C-index on BRCA, STAD, UCEC, COADREAD, and LIHC. In terms of average rank, LWR obtains 2.35, outperforming IntegrAO, JASMINE, and MSNE, and remaining close to MIND. This result suggests that LWR embeddings preserve clinically relevant biological variation beyond class-discriminative information.

The reconstruction results in Table~\ref{tab:reconstruction} further evaluate whether the learned representations preserve modality-level information. LWR achieves the highest Pearson correlation across all reported TCGA cohorts, as well as on CCMA and CCLE. These results indicate that the \rev{latent representations} learned by LWR retain substantial information about the original omics modalities, despite being constructed only from available \rev{modalities}.

Taken together, the results support the central motivation of LWR as an incomplete multi-omics representation learning method. Compared with graph-based \cite{ma2025moving} or network-embedding \cite{xu2021network} baselines, LWR benefits from neural latent fusion over heterogeneous omics modalities. Compared with JASMINE \cite{ballard2025jasmine}, LWR avoids relying on explicit missing-modality imputation as the basis for downstream evaluation. The empirical results show that this design yields strong and transferable embeddings, achieving the best average rank for cancer phenotype classification, competitive survival prediction performance, and the best reconstruction correlations across all reported datasets.
\subsection{Ablation Study}

\begin{table}[htb]
\centering
\small
\setlength{\tabcolsep}{3.5pt}
\resizebox{\textwidth}{!}{%
\begin{tabular}{lcccccc}
\toprule
Cancer type & LWR & Mean+Neighbor & Attn+Naive & Attn+None & Mean+Naive & Mean+None \\
\midrule
LUAD & 0.500 & \textbf{0.508} & 0.443 & 0.502 & 0.439 & 0.496 \\
KIRC & \textbf{0.557} & 0.503 & 0.454 & 0.533 & 0.447 & 0.525 \\
BRCA & \textbf{0.537} & 0.535 & 0.518 & 0.532 & 0.524 & 0.494 \\
LGG & -- & -- & -- & -- & -- & -- \\
OV & -- & -- & -- & -- & -- & -- \\
SKCM & 0.450 & \textbf{0.467} & 0.377 & 0.457 & 0.426 & 0.406 \\
THCA & 0.505 & \textbf{0.539} & 0.495 & 0.519 & 0.539 & 0.501 \\
BLCA & 0.468 & \textbf{0.490} & 0.398 & 0.463 & 0.298 & 0.471 \\
STAD & \textbf{0.456} & 0.387 & 0.389 & 0.422 & 0.403 & 0.396 \\
UCEC & -- & -- & -- & -- & -- & -- \\
COADREAD & 0.335 & \textbf{0.347} & 0.345 & 0.337 & 0.289 & 0.337 \\
COAD & 0.335 & 0.353 & 0.304 & 0.367 & \textbf{0.378} & 0.333 \\
GBM & -- & -- & -- & -- & -- & -- \\
HNSC & 0.555 & \textbf{0.559} & 0.504 & 0.553 & 0.450 & 0.546 \\
LUSC & 0.488 & 0.496 & 0.468 & 0.464 & 0.440 & \textbf{0.506} \\
LIHC & 0.464 & \textbf{0.493} & 0.425 & 0.456 & 0.462 & 0.470 \\
CESC & 0.480 & 0.450 & 0.437 & \textbf{0.483} & 0.453 & 0.380 \\
\midrule
Average & \textbf{0.472} & 0.471 & 0.428  & 0.468  & 0.427  & 0.451  \\
\bottomrule
\end{tabular}%
}
\caption{Five-fold cross-validation accuracy for cancer stage classification on \textbf{TCGA} datasets under different ablation settings. The full LWR model uses attention-based fusion with neighborhood-aware alignment. The best result on each dataset is highlighted in bold, and `--' indicates unavailable stage labels or skipped evaluation. The average is computed over datasets with valid results.}
\label{tab:tcga_stage_ablation}
\end{table}

\begin{table}[htb]
\centering
\small
\setlength{\tabcolsep}{3.5pt}
\resizebox{\textwidth}{!}{%
\begin{tabular}{lcccccc}
\toprule
Cancer type & LWR & Mean+Neighbor & Attn+Naive & Attn+None & Mean+Naive & Mean+None \\
\midrule
LUAD & 0.562 & 0.572 & 0.500 & 0.587 & 0.500 & \textbf{0.613} \\
KIRC & 0.706 & \textbf{0.712} & 0.595 & 0.705 & 0.595 & 0.709 \\
BRCA & \textbf{0.671} & 0.639 & 0.640 & 0.641 & 0.640 & 0.642 \\
LGG & 0.812 & 0.811 & 0.740 & 0.805 & 0.740 & \textbf{0.818} \\
OV & 0.598 & \textbf{0.612} & 0.600 & 0.594 & 0.600 & 0.604 \\
SKCM & 0.614 & \textbf{0.640} & 0.603 & 0.630 & 0.603 & 0.620 \\
THCA & 0.760 & 0.795 & \textbf{0.890} & 0.787 & \textbf{0.890} & 0.760 \\
BLCA & 0.615 & \textbf{0.620} & 0.598 & 0.615 & 0.598 & 0.612 \\
STAD & 0.563 & \textbf{0.607} & 0.531 & 0.541 & 0.531 & 0.574 \\
UCEC & 0.665 & 0.673 & 0.586 & 0.660 & 0.586 & \textbf{0.683} \\
COADREAD & \textbf{0.608} & 0.600 & 0.606 & 0.584 & 0.606 & 0.555 \\
COAD & 0.541 & 0.537 & 0.577 & \textbf{0.589} & 0.577 & 0.549 \\
GBM & 0.651 & \textbf{0.655} & \textbf{0.655} & \textbf{0.655} & \textbf{0.655} & \textbf{0.655} \\
HNSC & 0.577 & 0.585 & 0.548 & 0.582 & 0.548 & \textbf{0.598} \\
LUSC & 0.495 & 0.564 & 0.525 & \textbf{0.566} & 0.525 & 0.525 \\
LIHC & \textbf{0.639} & 0.631 & 0.518 & 0.591 & 0.521 & 0.564 \\
CESC & \textbf{0.651} & 0.622 & 0.553 & 0.611 & 0.553 & 0.595 \\ \midrule
Average & 0.631 & \textbf{0.640} & 0.604  & 0.632  & 0.604  & 0.628  \\
\bottomrule
\end{tabular}%
}
\caption{Five-fold cross-validation C-index for survival prediction on \textbf{TCGA} datasets under different ablation settings. The ablation study compares two fusion strategies, attention-based fusion and mean fusion, and three alignment strategies, neighborhood-aware alignment, naive pairwise alignment, and no alignment. The full LWR model corresponds to attention-based fusion with neighborhood-aware alignment. The best result on each dataset is highlighted in bold. The average is computed over all evaluated cancer types.}
\label{tab:tcga_survival_ablation}
\end{table}

\begin{table}[htb]
\centering
\small
\setlength{\tabcolsep}{3.5pt}
\begin{tabular}{lcccccc}
\toprule
Cancer type & LWR & Mean+Neighbor & Attn+Naive & Attn+None & Mean+Naive & Mean+None \\
\midrule
LUAD & 0.365 & 0.375 & -0.021 & 0.384 & -0.021 & \textbf{0.389} \\
KIRC & \textbf{0.475} & 0.472 & 0.016 & 0.461 & 0.017 & 0.457 \\
BRCA & 0.473 & 0.469 & -0.003 & 0.476 & -0.002 & \textbf{0.481} \\
LGG & 0.591 & 0.593 & -0.015 & 0.590 & -0.013 & \textbf{0.597} \\
OV & \textbf{0.331} & 0.323 & -0.013 & 0.327 & -0.017 & 0.329 \\
SKCM & 0.412 & 0.410 & -0.017 & 0.416 & -0.010 & \textbf{0.420} \\
THCA & \textbf{0.453} & 0.428 & 0.005 & 0.423 & -0.015 & 0.427 \\
BLCA & 0.436 & 0.441 & -0.031 & \textbf{0.447} & -0.028 & 0.435 \\
STAD & 0.436 & \textbf{0.452} & -0.012 & 0.448 & -0.010 & 0.445 \\
UCEC & 0.459 & 0.450 & -0.023 & \textbf{0.464} & -0.023 & 0.460 \\
COADREAD & \textbf{0.443} & 0.427 & -0.015 & 0.431 & -0.018 & 0.440 \\
COAD & 0.368 & 0.363 & -0.015 & 0.372 & -0.027 & \textbf{0.376} \\
GBM & \textbf{0.267} & 0.230 & -0.044 & 0.225 & -0.049 & 0.227 \\
HNSC & \textbf{0.399} & 0.392 & -0.034 & 0.385 & -0.033 & 0.399 \\
LUSC & 0.355 & 0.347 & -0.017 & 0.337 & -0.013 & \textbf{0.363} \\
LIHC & \textbf{0.527} & 0.519 & -0.022 & 0.514 & -0.024 & 0.518 \\
CESC & 0.332 & 0.328 & -0.039 & 0.321 & -0.054 & \textbf{0.343} \\
\midrule
Overall & \textbf{0.419} & 0.413 & -0.018 & 0.413 & -0.020 & 0.418 \\
\bottomrule
\end{tabular}
\caption{TCGA masked reconstruction performance by cancer type. Each entry is the mean Pearson correlation across available modalities for that cancer type. The best ablation setting per row is bolded.}
\label{tab:tcga_masked_reconstruction_by_cancer_ablation}
\end{table}
To systematically evaluate the contributions of the two core components in LWR—\textbf{attention-based fusion} and \textbf{neighborhood-aware alignment}—we design an ablation study around their replaceable variants. Specifically, we evaluate the fusion module by comparing adaptive attention against unweighted mean fusion, and we assess the alignment module by comparing neighborhood-aware topology preservation against both naive pairwise alignment and a no-alignment baseline.

\paragraph{Ablated variants.} We define three standard alternatives to isolate the effects of our proposed modules:
\begin{itemize}
    \item \textbf{Ablating adaptive weighting (Mean fusion)}: To evaluate the necessity of the attention module, we replace it with simple mean fusion, which aggregates the available modalities $\mathcal{A}_i$ uniformly: $z_i^{\mathrm{mean}} = \frac{1}{|\mathcal{A}_i|} \sum_{m \in \mathcal{A}_i} z_i^m$. While simple and naturally robust to missing \rev{modalities}, it assumes an identical signal-to-noise ratio across all modalities. This uniform weighting inherently risks the dilution of highly informative, modality-specific biological features with uninformative or noisy \rev{modalities}.
    \item \textbf{Ablating topology preservation (Pairwise alignment)}: To assess the effectiveness of neighborhood-aware alignment, we replace it with naive pairwise alignment. This variant enforces strict coordinate-wise equivalence by minimizing the squared Euclidean distance between paired representations: $\mathcal{L}_{\mathrm{naive}} = \frac{1}{|\mathcal{P}|} \sum_{(m,n) \in \mathcal{P}} \| z_i^m - z_i^n \|_2^2$. This constraint can over-regularize modality-specific representations by forcing different \rev{modalities} of the same sample to occupy nearby coordinates, potentially suppressing complementary modality-specific signals.
    \item \textbf{Ablating alignment entirely (No alignment)}: To test whether latent space regularization is necessary at all, we remove the alignment loss completely. In this configuration, the modality-specific encoders project data into unconstrained latent spaces, relying entirely on the fusion module to bridge the modalities. This lack of explicit harmonization can lead to disjoint representations that fail to capture shared biological structures.
\end{itemize}
We evaluate the complete $2 \times 3$ factorial combinations of these fusion and alignment strategies across three downstream tasks: cancer stage classification (Table \ref{tab:tcga_stage_ablation}), patient survival prediction (Table \ref{tab:tcga_survival_ablation}), and masked modality reconstruction (Table \ref{tab:tcga_masked_reconstruction_by_cancer_ablation}).

\paragraph{Impact of alignment strategies.} The empirical results demonstrate that naive pairwise alignment causes significant representation degradation across all tasks. Most notably, as shown in Table \ref{tab:tcga_masked_reconstruction_by_cancer_ablation}, configurations employing naive alignment (Attn+Naive and Mean+Naive) lead to near-collapse in reconstruction quality, yielding overall correlations near zero ($-0.018$ and $-0.020$). This representation collapse translates to severely degraded performance in classification (Table \ref{tab:tcga_stage_ablation}). The central takeaway from these comparisons is that neighbor-based alignment successfully avoids such collapse induced by strict coordinate-wise constraints. While its explicit benefit over unaligned baselines is more modest in certain tasks like survival prediction, it acts as a critical structural safeguard that preserves modality-specific relational geometry without over-regularizing the latent space.

\paragraph{Impact of fusion strategies.} The comparative performance of attention-based versus mean fusion exhibits clear task-dependency, highlighting a fundamental trade-off between dynamic feature selection and regularization. For cancer stage classification (Table \ref{tab:tcga_stage_ablation}), the full LWR model (Attention + Neighborhood) achieves the strongest overall performance (average accuracy 0.472), indicating that dynamic weighting is effective when specific omics \rev{modalities} carry distinct diagnostic relevance. For masked reconstruction (Table \ref{tab:tcga_masked_reconstruction_by_cancer_ablation}), performance among stable variants is highly competitive, though the full LWR model retains a marginal edge (0.419). Conversely, for patient survival prediction, Table \ref{tab:tcga_survival_ablation} reveals that the Mean+Neighbor configuration achieves a higher average C-index (0.640) compared to the full LWR model (0.631). In such complex scenarios, the uniform weighting of mean fusion acts as a robust structural regularizer, mitigating the risk of overfitting to the dynamic attention weights of specific \rev{modalities}.

\subsection{Case Study} 
\subsubsection{Clustering-Based Survival Stratification}
To further examine the clinical interpretability of the learned representations, we conduct a clustering-based case study on TCGA cohorts. Unlike the supervised survival prediction experiment, this analysis directly investigates whether the latent space learned by LWR can naturally organize patients into subgroups with distinct survival outcomes.

For each cancer type, we apply clustering to the learned patient representations and vary the number of clusters from $k=3$ to $k=8$. We then evaluate the survival difference among the resulting patient groups using the log-rank test. The clustering quality is measured by $-\log_2(p)$, where p is the log-rank test p-value. A larger value indicates stronger survival separation, suggesting that the latent representation captures more survival-relevant patient heterogeneity.

\begin{figure}
    \centering
    \includegraphics[width=0.8\textwidth]{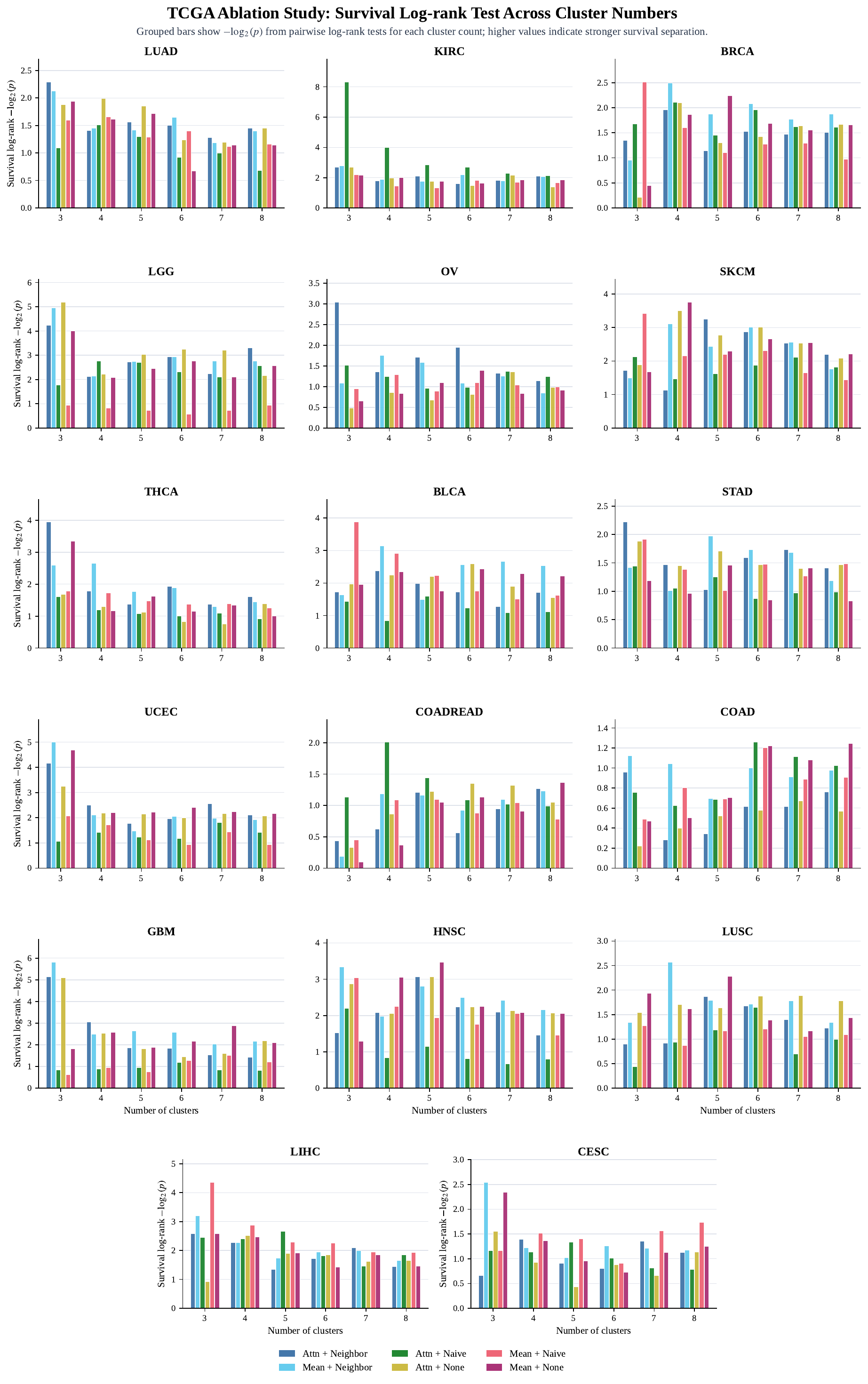}
\caption{Case study of clustering-based survival stratification on \textbf{TCGA} datasets. For each cancer type, patients are clustered using learned latent representations with \(k \in \{3,4,5,6,7,8\}\). Survival differences among clusters are evaluated using the log-rank test, and bar heights indicate \(-\log_2(p)\), where higher values suggest stronger survival separation. Simplified LWR variants are included as reference models for comparison.}
\label{fig:tcga_ablation_clustering}
\end{figure}

Figure~\ref{fig:tcga_ablation_clustering} presents the clustering-based survival stratification results across TCGA cancer types. In this case study, we also include several simplified variants of LWR as reference models, including variants with mean fusion, naive alignment, or no alignment.

The results show that LWR generally produces competitive and stable survival separation across different cancer types and cluster numbers. This suggests that the representations learned by LWR contain clinically meaningful structure beyond what is captured by standard downstream prediction metrics. In several cohorts, the patient clusters obtained from LWR exhibit clear survival differences, indicating that the latent space can reflect underlying disease heterogeneity.

Meanwhile, the comparison with simplified variants shows that the structure of the latent space is sensitive to both fusion and alignment choices. Variants based on naive pairwise alignment or no alignment sometimes produce strong separation for specific cancer types or cluster numbers, but their performance tends to be less consistent across cohorts. This observation suggests that directly forcing modality-specific embeddings to be close, or removing alignment entirely, may not reliably preserve clinically relevant patient relationships. By contrast, LWR uses neighborhood-aware alignment to preserve local sample structures while avoiding overly restrictive point-wise alignment, which may help maintain a more meaningful organization of patients in the fused latent space.

\subsection{Clinical and Biological Interpretability}
\begin{figure}[tp]
    \centering
    
    % (A) Mortality rates
    \begin{subfigure}[b]{0.6\textwidth}
        \centering
        \includegraphics[width=\textwidth]{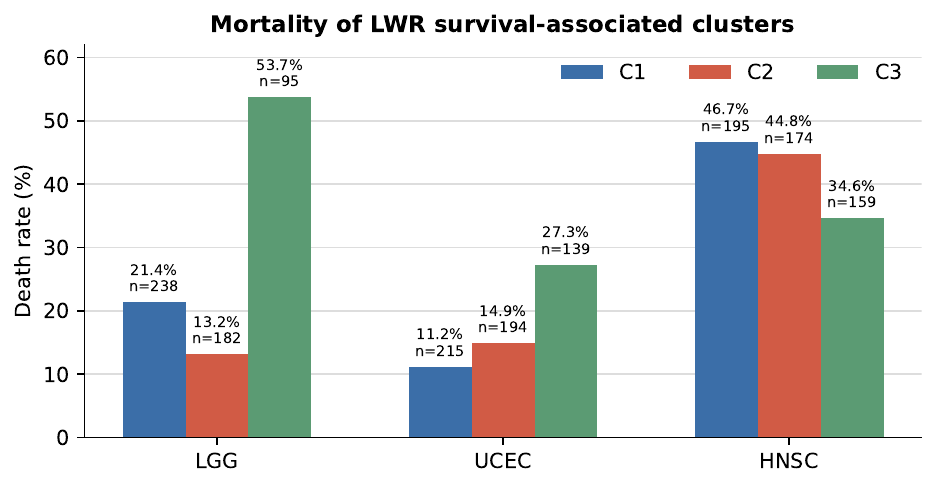}
        \caption{Mortality rates across the identified patient clusters.}
        \label{fig:bio_clusters_a}
    \end{subfigure}
    
    \vspace{1.5em} % 控制子图之间的垂直间距
    
    % (B) Cluster composition
    \begin{subfigure}[b]{0.6\textwidth}
        \centering
        \includegraphics[width=\textwidth]{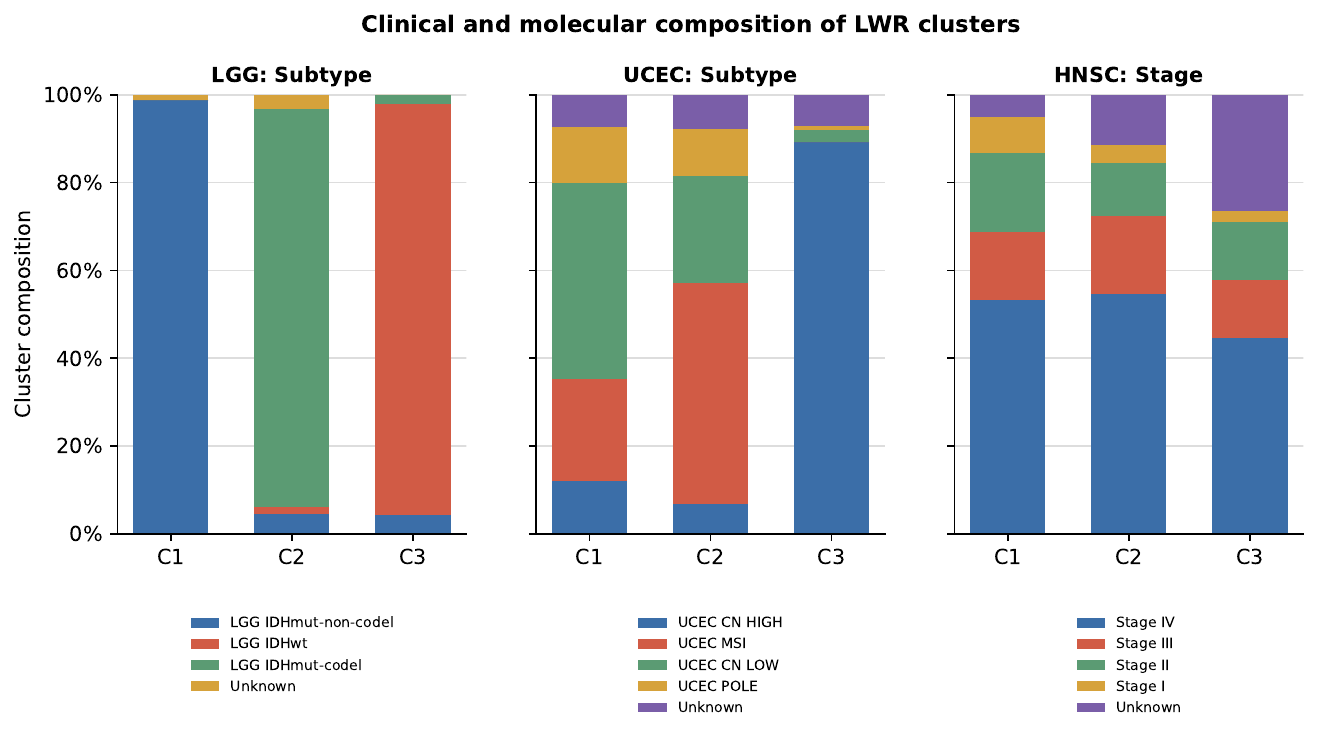}
        \caption{Clinical and molecular composition of the clusters.}
        \label{fig:bio_clusters_b}
    \end{subfigure}
    
    \vspace{1.5em} % 控制子图之间的垂直间距
    
    % (C) Modality attention
    \begin{subfigure}[b]{0.6\textwidth}
        \centering
        \includegraphics[width=\textwidth]{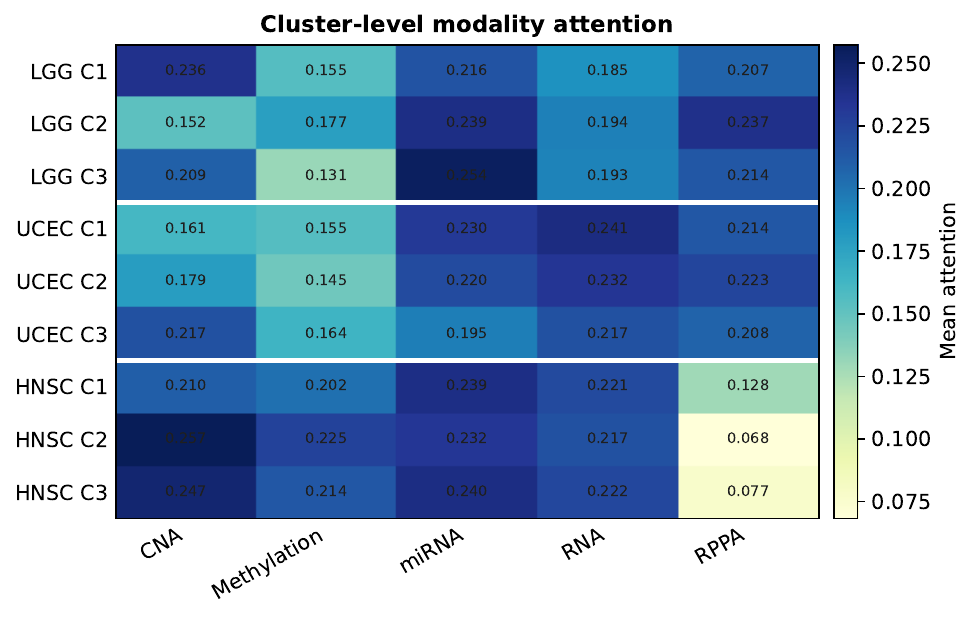}
        \caption{Cluster-level average modality attention weights.}
        \label{fig:bio_clusters_c}
    \end{subfigure}
    
    \caption{Biological and clinical characterization of LWR survival-associated clusters ($k=3$) in LGG, UCEC, and HNSC. \textbf{(a)} Mortality rates showing survival stratification across different patient groups. \textbf{(b)} Representation enrichment demonstrating the alignment with specific prognostic subtypes. \textbf{(c)} Model's dynamic reliance on different omics modalities, illustrating availability-aware modality fusion.}
    \label{fig:bio_clusters}
\end{figure}

To further strengthen the biological interpretation of the survival-associated clusters, we characterize patient subgroups from three representative cancer types that exhibited strong survival stratification signals: LGG, UCEC, and HNSC (Figure~\ref{fig:bio_clusters_a}). For each cohort, we partitioned the LWR latent representations into \(k=3\) clusters and annotated the resulting groups using available clinical and molecular variables, including molecular subtype, tumor stage, and survival data (Table~\ref{tab:biological_cluster_characterization}).

The clearest biological alignment is observed in LGG, where the learned clusters closely capture established glioma molecular subtypes (Figure~\ref{fig:bio_clusters_b}). Cluster 1 is highly enriched for IDH-mutant/non-codeleted tumors (98.7\% of the cluster) and exhibits an intermediate mortality rate (21.4\%). Cluster 2 is dominated by IDH-mutant/1p19q-codeleted tumors (90.7\%) and shows the lowest mortality (13.2\%). In sharp contrast, Cluster 3 is strongly enriched for IDH-wildtype tumors (93.7\%) and presents the highest mortality rate (53.7\%). This stratification perfectly aligns with the known poor prognosis of IDH-wildtype gliomas, demonstrating that the unsupervised LWR latent space successfully recovers highly prognostic, biologically meaningful disease heterogeneity.

A similarly interpretable pattern emerges in UCEC. The highest-risk group (Cluster 3, mortality 27.3\%) is predominantly composed of copy-number-high (CN-high) tumors (89.2\%), which are characteristically dominated by aggressive serous endometrial adenocarcinomas. Conversely, Clusters 1 and 2 are enriched for CN-low and microsatellite instability (MSI) tumors, respectively, corresponding to their more favorable clinical outcomes. In HNSC, while the subtype enrichment is more modest, the clusters still display distinct mortality and stage profiles, suggesting the representations capture multi-faceted patient heterogeneity beyond a single clinical axis.

Furthermore, we examine the cluster-level modality-attention weights to understand how LWR's fusion module dynamically prioritizes different omics layers (Figure~\ref{fig:bio_clusters_c}). The attention patterns are remarkably cancer-specific. In LGG, the IDH-mutant/non-codeleted cluster heavily relies on copy-number alterations (CNA), whereas the IDH-wildtype and IDH-mutant/codeleted clusters assign higher attention to miRNA. In UCEC, RNA-seq universally receives the highest attention, though the aggressive CN-high cluster naturally exhibits elevated attention to CNA. In HNSC, miRNA and CNA emerge as the dominant modalities. These quantitative variations confirm that the availability-aware fusion mechanism does not default to a static averaging strategy; rather, it adaptively integrates omics sources based on the underlying biological architecture of each cancer cohort. 

Overall, this analysis demonstrates that the survival-associated clusters are not merely abstract numerical groupings. Instead, they correspond to known molecular subtypes, histological patterns, and clinically distinct patient profiles. These findings strongly validate the biological interpretability of the LWR representations and bridge the gap between missing-modality latent learning and established cancer biology.

\begin{table}[htb]
\centering
\resizebox{\textwidth}{!}{
\begin{tabular}{llrrlllc}
\toprule
Cancer & Cluster & $n$ & Death rate & Median time (days) & Dominant annotation & Proportion & Top modality \\
\midrule
LGG & 1 & 238 & 21.4\% & 842 & LGG IDHmut-non-codel & 98.7\% & CNA \\
LGG & 2 & 182 & 13.2\% & 706 & LGG IDHmut-codel & 90.7\% & miRNA \\
LGG & 3 & 95 & 53.7\% & 491 & LGG IDHwt & 93.7\% & miRNA \\
\midrule
UCEC & 1 & 215 & 11.2\% & 945 & UCEC CN LOW & 44.7\% & RNA \\
UCEC & 2 & 194 & 14.9\% & 944 & UCEC MSI & 50.5\% & RNA \\
UCEC & 3 & 139 & 27.3\% & 832 & UCEC CN HIGH & 89.2\% & RNA \\
\midrule
HNSC & 1 & 195 & 46.7\% & 625 & Stage IV & 53.3\% & miRNA \\
HNSC & 2 & 174 & 44.8\% & 572 & Stage IV & 54.6\% & CNA \\
HNSC & 3 & 159 & 34.6\% & 783 & Stage IV & 44.7\% & CNA \\
\bottomrule
\end{tabular}
}
\caption{Quantitative biological characterization of LWR survival-associated clusters in selected TCGA cohorts ($k=3$).}
\label{tab:biological_cluster_characterization}
\end{table}

\section{Conclusion}
We presented \emph{Latent World Recovery} (LWR), a VAE-based framework for incomplete \rev{multimodal representation learning}. Instead of treating missing modalities as observations that must be synthesized before downstream analysis, LWR learns sample-level representations directly from the \rev{modalities} that are available. The framework combines \rev{modality-specific variational encoders}, \rev{availability-aware modality fusion}, \rev{observed-modality reconstruction}, and neighbor-based latent alignment. This design allows LWR to preserve modality-level information while organizing heterogeneous \rev{modalities} in a shared latent space without enforcing overly restrictive coordinate-level agreement.

Experiments on incomplete multi-omics benchmarks demonstrate that LWR learns effective representations across cancer phenotype classification, survival prediction, and reconstruction-based information preservation. The ablation results further show that the choice of alignment strategy is critical: naive pairwise alignment can substantially degrade representation quality, whereas neighbor-based alignment provides a more stable way to transfer local sample structure from individual modalities to the fused representation. The clustering-based survival stratification case study also suggests that the learned latent space can capture clinically meaningful patient heterogeneity beyond standard supervised evaluation metrics.

Overall, our results support the view that incomplete \rev{multimodal learning} should focus less on reconstructing absent modalities and more on recovering robust representations from partial observations. LWR offers a practical step in this direction by separating \rev{missing-modality representation learning} from task-specific prediction and by enabling downstream models to operate on fused embeddings constructed only from observed \rev{modalities}. Future work will investigate more expressive uncertainty-aware fusion mechanisms, extend the framework to larger and more diverse biomedical datasets, and study how latent neighborhood alignment can be further adapted to modality-specific biological structure.
\bibliographystyle{plain}
\bibliography{references}

\end{document}